\documentclass{article}
\usepackage{spconf,amsmath,graphicx,url}
 
\title{AUTOMATED PULMONARY NODULE DETECTION USING \\3D DEEP CONVOLUTIONAL NEURAL NETWORKS}

\name{Hao Tang\sthanks{These authors have contributed equally to this work.}$^{\dagger}$, Daniel R. Kim\footnotemark[1]$^{\ddagger}$, Xiaohui Xie$^{\dagger}$}

\address{$^{\dagger}$Department of Computer Science, University of California, Irvine \\$^{\ddagger}$University of California, Irvine School of Medicine}
\begin{document}
%
\maketitle
\begin{abstract}

Early detection of pulmonary nodules in computed tomography (CT) images is essential for successful outcomes among lung cancer patients. Much attention has been given to deep convolutional neural network (DCNN)-based approaches to this task, but models have relied at least partly on 2D or 2.5D components for inherently 3D data. In this paper, we introduce a novel DCNN approach, consisting of two stages, that is fully
three-dimensional end-to-end and utilizes the state-of-the-art in object detection. First, nodule candidates are identified with a U-Net-inspired 3D Faster R-CNN trained
using online hard negative mining. Second, false positive reduction is
performed by 3D DCNN classifiers trained on difficult examples produced during candidate screening. Finally, we introduce a method to ensemble models from both stages via consensus to give the final predictions. By using this framework, we ranked first of 2887 teams in Season One of Alibaba's 2017 TianChi AI Competition for Healthcare.
 
\end{abstract}
\section{Introduction}
\label{sec:intro}
 
Lung cancer has been the leading cause of all cancer-related deaths, causing 1.3 millions death annually \cite{Siegel15}. Detecting pulmonary nodules early is critical for a good prognosis of the disease, and low-dose computed tomography (CT) scans are widely used and very effective for this purpose. However, manually screening CT images is time-consuming for radiologists who are increasingly overwhelmed with data. Advanced computer-aided diagnosis systems (CADs) have the potential to expedite
this process but the task is complicated by the variation in nodule size (from 3 to 50 mm), shape, density,
and
anatomical context, as well as the abundance of tissues that resemble the appearance of nodules (e.g., blood vessels, chest wall). 

Many approaches have been proposed for this challenge, often employing two stages: 1) nodule candidate screening, which identifies candidates with high sensitivity at the expense of accumulating many false positives, and 2) false positive reduction. Frameworks for the first stage commonly relied on techniques including voxel clustering and curvature computation \cite{Jacobs14,Murphy09}, while second stage methods carefully utilized low-level descriptors such as intensity, size, sphericity, texture, and contextual information \cite{Jacobs14,Murphy09,Ginneken10}. These conventional methods had limited discriminative power due to their reliance on hand-crafted features. More recent efforts have focused on the use of convolutional neural networks (CNNs) and have produced encouraging
results, but often use 2D or 2.5D networks in some components for inherently 3D data \cite{Ding17,Setio16,Luna16}; nodules can be impossible to discriminate from tissues such as blood vessels from axial slices. Moreover, frameworks incorporating the state-of-the-art models in object detection are still rare.

In this work, we propose a novel CAD framework that consists entirely of three-dimensional deep convolutional neural networks (3D DCNNs) end-to-end. Candidate detection is first performed by a U-Net \cite{Ronneberger15}-like Faster Region-based CNN (Faster R-CNN) \cite{Ren15}, which is the state-of-the-art model in object detection. The hard mimics identified by
the detector
are then used to train highly discriminative, deep 3D classifiers for
false positive reduction. Both models heavily utilize residual shortcuts \cite{He15} that promote performance gains with deep architectures. The final prediction scores are generated by ensembling the detector and the classifiers, unifying contributions learned from both stages. We validated our proposed method on a dataset of 1000 low-dose CT images provided by the 2017 TianChi AI Competition for Healthcare organized by Alibaba \cite{TianChi17}, where our model ranked 1st in Season One. 
 
\section{RELATED WORK}
\label{sec:work}
 
Ding et al. (2017) proposed a CAD system using Faster R-CNN on 2D axial slices then false positive reduction with a 3D DCNN \cite{Ding17}. Dou et al. (2017) leverages 3D input but uses a binary classifier 3D CNN with online sample filtering for candidate screening, rather than a Faster R-CNN \cite{Dou17}. Our work utilizes an efficient 3D Faster R-CNN for detection and deep residual 3D classifiers for false positive reduction.

\section{PROPOSED FRAMEWORK}
\label{sec:framework}

\begin{figure*}[htb]
  \centerline{\includegraphics[width=14cm]{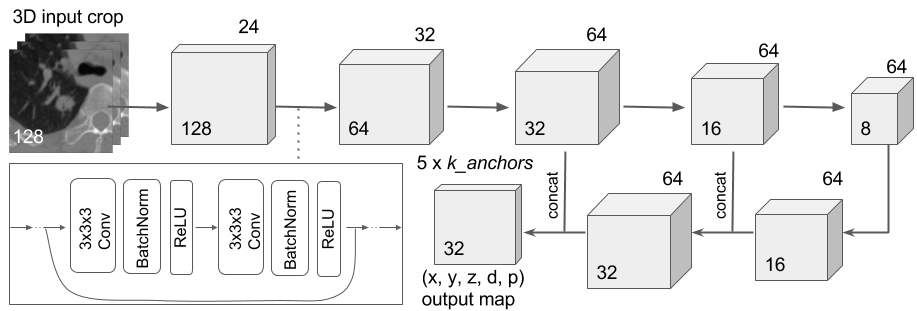}}
\caption{General architecture of the candidate-screening 3D Faster R-CNN.}
\label{fig:detector}
\end{figure*} 

Our proposed method for nodule detection roughly follows two stages: (1) candidate screening using a 3D Faster R-CNN, and (2) subsequent false positive reduction using 3D DCNN classifiers. The purpose of the Faster R-CNN in (1) is to identify nodule candidates while preserving high sensitivity, whereas the classifiers in (2) finely discriminate between true nodules and false positives. We find optimal results when models from both stages are ensembled
for final predictions.

Rather than using one stage in which we heavily retrain the Faster R-CNN with hard examples, we believe the two-stage framework provides more flexiblity in adjusting the trade-off between sensitivity and specificity.
 
\subsection{Candidate Screening Using 3D Faster R-CNN}
\label{ssec:detector}

The success of Faster R-CNN \cite{Ren15} and deep residual networks \cite{He15} in natural images, and U-Net \cite{Ronneberger15} in medical images, has inspired the use of a deep residual 3D Faster R-CNN architecture with transposed convolutional layers, which is illustrated in Fig. \ref{fig:detector}. After a series of downsampling layers to encode high-level information, we concatenate early features with latter ones and feed them through several upsampling transposed convolutions, decoding
high-resolution information regarding the nodule's location and diameter. Because we use over 30 convolutional layers, we use residual shortcuts extensively. Memory
limits on 4 GPUs made it necessary to split the input image into overlapping $128 \times 128 \times 128$ input volumes, process them separately, and combine them.
 
The output is a $32 \times 32 \times 32$ map of $(x, y, z)$ coordinates, diameter, and nodule probability corresponding to regions of the input volume. These five features are parameterized by three anchors whose sizes we set to 5, 10, and 30 mm based on the nodule size distribution in our dataset. Each input region is associated with output for each anchor, so the output map is of shape $32 \times 32 \times 32 \times 5 \times 3$. 

We compute a classification loss $L_{cls}$ for the predicted nodule probabilities and four regression losses $L_{reg}$ associated with predicted nodule coordinates and diameters. The ground truth labels are determined for each anchor as follows. If an anchor $i$ overlaps with a nodule with an intersection over union (IoU) equal or greater than a threshold of 0.5, we regard it as positive $(p_i^*=1)$. In contrast, if anchor $i$ overlaps with a
nodule with an IoU less than 0.2, we regard it as negative
$(p_i^*=0)$. All other anchors do not contribute to the loss. Note also that only positive anchors contribute to the regression loss. The final loss for anchor $i$ is defined as
 
\begin{equation}
L(p_i, t_i) = \lambda L_{cls}(p_i, p_i^*)+p_i^*L_{reg}(t_i, t_i^*)
\end{equation}
 
where $p_i$ is the predicted nodule probability; $t_i$ is the vector  
\begin{equation}
t_i=\left(\frac{x-x_a}{d_a}, \frac{y-y_a}{d_a}, \frac{z-z_a}{d_a}, \log\frac{d}{d_a}\right)
\end{equation}
of predicted relative coordinates and diameter, where $x, y, z, d$ are the predicted nodule coordinates and diameter and \\$x_a, y_a, z_a, d_a$ are the coordinates and size of anchor $i$. Similarly, the ground truth nodule is expressed as the vector
 
\begin{equation}
t_i^*=\left(\frac{x^*-x_a}{d_a}, \frac{y^*-y_a}{d_a}, \frac{z^*-z_a}{d_a}, \log\frac{d^*}{d_a}\right)
\end{equation}
 
where $x^*, y^*, z^*, d^*$ are the coordinates and diameter of the ground truth box. We set $\lambda$ to 1. We use binary cross entropy loss for $L_{cls}$ and smooth $L1$ loss for $L_{reg}$.

\subsubsection{Hard Negative Mining}
\label{sssec:mining}

Each input volume to the Faster R-CNN is dominated by numerous trivial negative locations (air). To make the negative samples as informative as possible, we used hard negative mining \cite{Shrivastava16}. A pool of $N$ predictions corresponding to condition negative anchors were randomly selected and ranked in descending order according to nodule probability. The top $n$ samples were chosen to be considered in the loss function.

\subsection{False Positive Reduction Using 3D DCNN Classifier}
\label{ssec:classifier}

\begin{figure}[htb]
  \centerline{\includegraphics[width=4.5cm]{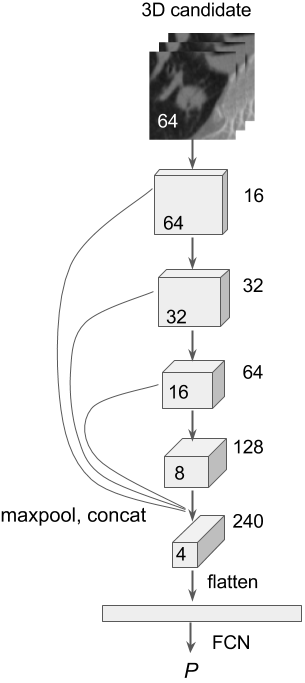}}
\caption{General architecture of the 3D false positive reduction classifier.}
\label{fig:classifier}
\end{figure} 

The predictions $(x, y, z, d, p)$ from the Faster R-CNN are used to extract $64 \times 64 \times 64$ crops centered at $(x, y, z)$ for input to a DCNN classifier, whose architecture is illustrated in Fig. \ref{fig:classifier}. It begins with several residual blocks of Conv, BatchNorm, ReLU layers, which are ultimately fed to a fully-connected layer to calculate the final classification score. We integrate detailed local information about the nodule with more contextual features
by adding shortcuts from the end of each block to the last feature map.
 
\section{EXPERIMENTS AND RESULTS}
\label{sec:results}

We validated our framework on the large-scale TianChi competition dataset, which contains CT scans from 1000 patients from hospitals in China. The images were annotated by radiologists similarly to LUNA16 \cite{Luna16}, i.e. with nodule location and size. We used 600 images for training (containing 969 annotated nodules), 200 for validation, and the remaining 200 comprised the test set.

The evaluation metrics included sensitivity and average number of false positives per scan (FPs/scan), where a detection is considered a true positive if the location falls within the radius of a nodule centroid. The competition ranked participants based on a CPM score defined as the average sensitivity at seven predefined FPs/scan rates: 1/8, 1/4, 1/2, 1, 2, 4, 8.

\subsection{Training}
\label{ssec:training}

The Faster R-CNN was trained with Adam for 150 epochs. The examples for each epoch were split such that 70\% of the examples consisted of the entire training set of annotations (positive samples), and 30\% consisted of random nodule-lacking cropped images from random scans (negative samples). The classifier was trained for 300 epochs with Adam using the same positive examples as the Faster R-CNN detector. These were balanced with hard negative samples, i.e. 969 samples for which the detector
assigned a confidence score of 0.5 or greater. The input candidates for test set predictions were provided by the detector. For both detector and classifier, the checkpoint with the highest CPM on the validation set was used for prediction on the test set. 

\subsection{Data Augmentation}
\label{ssec:data_aug}
We trained the Faster R-CNN with random x-, y-, and z-axis flips; random scaling; and large jitters to promote translational invariance and improve generalization. The classifiers were trained similarly along with random rotations. Interestingly, even though the nodule locations predicted by the detector are expected to be centers, minor regression errors made it necessary to add small jitters of up to 2 mm.

\subsection{Faster R-CNN and Classifier Ensemble Results}
\label{ssec:ensemble}

\begin{table}
  \renewcommand{\arraystretch}{1.2}
  \caption{Validation score comparison showing stepwise performance gains with hard negative mining and classifier ensembling.}
  \begin{tabular}{ l r }
    \hline
    Prediction Method & Validation CPM \\
    \hline
    Faster R-CNN & 0.603 \\
    Faster R-CNN w/ hard negative mining & 0.695 \\
    Average (Faster R-CNN, classifier) & 0.723\\
    Consensus (Faster R-CNN, 3 classifers) & 0.758\\ 
    \hline\hline
  \end{tabular}
  \label{tab:cpm}
\end{table}

The stepwise performance gains of the Faster R-CNN with hard negative mining and classifier ensembles are shown in Table \ref{tab:cpm}. Hard negative mining substantially increased the validation CPM from 0.603 to 0.695, demonstrating the importance of using the most informative negative samples. The ensemble average of the Faster R-CNN and the classifier achieved an improved validation CPM of 0.723. 

\begin{figure}[htb]
  \includegraphics[width=8.5cm]{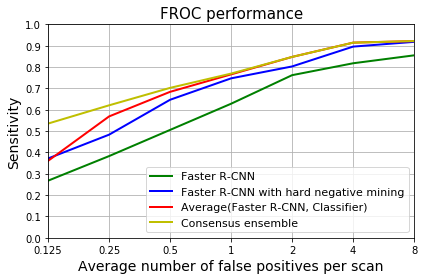}
\caption{Free-response receiver operating characteristic (FROC) curves showing stepwise performance gains in validation with hard negative mining and classifier ensembling.}
\label{fig:froc}
\end{figure}

The validation CPM was increased further still, to 0.758, with a “consensus ensembling” method that worked as follows. Two additional classifiers with similar architecture were trained. If the three classifiers agreed with the ensemble
average of the detector with the
first classifier that a particular candidate location was the most probable nodule for that patient, then the probability score was increased such that “consensus” candidates would rank higher than non-consensus ones across all test patients. Fig. \ref{fig:froc} presents the free-response receiver operating characteristic (FROC) curves. Note the increase in sensitivity at 0.125 and 0.25 FPs/scan.

Ultimately, this consensus ensembling method was used in calculating our final test set predictions for the competition. The CPM score was 0.815, surpassing all other submissions for the TianChi challenge. Final rankings are shown in Table \ref{tab:ranking}.

\begin{table}
  \renewcommand{\arraystretch}{1.2}
  \caption{Top 5 submissions to Season One of the TianChi challenge, for which 2887 teams participated.}
  \begin{tabular}{ l r }
    \hline
    Team & Test CPM \\
    \hline
    Ours & 0.815 \\
    Yi Yuan Smart HKBU & 0.806\\
    LAB2112 (qfpxfd) & 0.780\\ 
    Biana Information Technology & 0.780\\
    LAB518-CreedAI & 0.769\\
    \hline\hline
  \end{tabular}
  \label{tab:ranking}
\end{table}
 
\section{CONCLUSION}
\label{sec:conclusion}
 
In this paper, we present our fully three-dimensional framework of automatic pulmonary nodule detection. It consists of a U-Net-like 3D Faster R-CNN, trained with online hard negative mining, and a 3D classifier for false positive reduction. We introduce a consensus ensembling method to integrate both models for predictions. We validate our method  in the 2017 TianChi Healthcare AI Competition, achieving superior performance (0.815 CPM). We believe our model is a powerful clinical tool that harnesses state-of-the-art architectures in a way that captures the spatial nature of CT data.
 
\bibliographystyle{IEEEbib}
\bibliography{refs}
 
\end{document}